\documentclass[conference]{IEEEtran}
\IEEEoverridecommandlockouts
\usepackage{array}
\usepackage{makecell}
\usepackage{listings}
\usepackage{cite} 
\usepackage{amsmath,amssymb,amsfonts} 
\usepackage{algorithmic} 
\usepackage{graphicx} 
\usepackage{textcomp} 
\usepackage{xcolor} 
\usepackage[nolist]{acronym}
\usepackage{booktabs}
\usepackage{multirow}
\usepackage{etoolbox}

\def\BibTeX{{\rm B\kern-.05em{\sc i\kern-.025em b}\kern-.08em
    T\kern-.1667em\lower.7ex\hbox{E}\kern-.125emX}}

\begin{acronym}
  \acro{IoT}{Internet of Things}
  \acro{LLM}{Large Language Model}
  \acro{LLMs}{Large Language Models}
  \acro{IoBT}{Internet of Battlefield Things}
  \acro{QL}{Query Language}
  \acro{SQL}{Structured Query Language}
  \acro{NoSQL}{``Not only'' Structured Query Language}
  \acro{CQL}{Cypher Query Language}
  \acro{EM}{Exact Match}
  \acro{MSA}{Multi-Purpose Sensing Area}
\end{acronym}

\begin{document}

\title{

Natural Language Interaction with Databases on Edge Devices in the Internet of Battlefield Things\\

\thanks{Christopher Molek and K. Brent Venable were supported by NSF award NSF-RI-2008011.}
}

\author{
    \IEEEauthorblockN{
        Christopher D. Molek\IEEEauthorrefmark{1}\IEEEauthorrefmark{2},
        Roberto Fronteddu\IEEEauthorrefmark{2},
        K. Brent Venable\IEEEauthorrefmark{1}\IEEEauthorrefmark{2},
        Niranjan Suri\IEEEauthorrefmark{1}\IEEEauthorrefmark{2}\IEEEauthorrefmark{3}\\
    }
    
    \IEEEauthorblockA{
        \IEEEauthorrefmark{1} \textit{Department of Intelligent Systems \& Robotics - The University of West Florida (UWF), Pensacola, FL, USA}\\
        \IEEEauthorrefmark{2}\textit{Florida Institute for Human and Machine Cognition (IHMC), Pensacola, FL, USA}\\
        \textit{\{cmolek, rfronteddu, bvenable, nsuri\}@ihmc.org}\\
        \IEEEauthorrefmark{3}\textit{US Army DEVCOM Army Research Laboratory (ARL), Adelphi, MD, USA} \\
        \textit{niranjan.suri.civ@mail.mil}\\
    }
}

\maketitle

\begin{abstract}
The expansion of the \ac{IoT} in the battlefield, \ac{IoBT}, gives rise to new opportunities for enhancing situational awareness. To increase the potential of \ac{IoBT} for situational awareness in critical decision making, the data from these devices must be processed into consumer-ready information objects, and made available to consumers on demand. 
To address this challenge we propose a workflow that makes use of natural language processing (NLP) to query a database technology and return a response in natural language. Our solution utilizes \ac{LLMs}  that are sized for edge devices to perform NLP as well as graphical databases which are well suited for dynamic connected networks which are pervasive in the \ac{IoBT}. Our architecture employs LLMs for both mapping questions in natural language to Cypher database queries as well as  to summarize the database output back to the user in natural language. We evaluate several medium sized LLMs for both of these tasks on a database representing publicly available data from the US Army's Multi-purpose Sensing Area (MSA) at the Jornada Range in Las Cruces, NM. We observe that Llama 3.1 (8 billion parameters) outperforms the other models across all the considered metrics. Most importantly, we note that, unlike current methods,  our two step approach allows the relaxation of the \ac{EM} requirement of the produced Cypher queries with ground truth code and, in this way,  it 
achieves a 19.4\% increase in accuracy. Our workflow lays the ground work for deploying LLMs on edge devices to enable natural language interactions with databases containing  information objects for critical decision making.

\end{abstract}


\section{Introduction}
There is an increasingly significant body of research with a focus on \ac{IoT} - a network of interconnected devices ranging from temperature sensors  to mobile devices. 
Their ever increasing numbers are generating vast datasets requiring robust storage, processing, and analysis. Insights derived from this data enhance situational  awareness about the environment and support improved decision making\cite{tsiolisSmartCityAssisted2025}.
Real-time data processing and analytics are critical in military environments, particularly for \ac{IoBT} networks, where intermittent or limited connectivity depreciates access to cloud computing capabilities. Edge devices address this limitation by locally processing and analyzing data for utilization for the consumer\cite{bhardwajSurveyIntegrationOptimization2024}.

A notable capability of \ac{LLMs} in other research areas is their effectiveness in distilling useful information from extensive datasets \cite{luoLargeLanguageModels2025, bhardwajSurveyIntegrationOptimization2024, zhuLargeLanguageModels2024}. \ac{LLMs} have also been utilized to generate code specific outputs based on natural language questions\cite{wangReviewCodeGeneration2023}.  This is particularly useful, when considering the task of generating database queries from natural language prompts. 

This work aims at establishing both the capabilities and limitations of LLMs for on-device data extraction from large IoBT-generated corpora. Initially, we probe the structure of a simulated IoBT using graph databases, which outperform SQL databases in this context. Graph databases enable dynamic updates as nodes join or leave the network and efficiently manage deeply connected, multi-relational data. Their scalability allows queries to target only relevant graph segments, whereas SQL databases suffer from performance degradation due to costly join operations as the data volume increases or relationships are added. Additionally, graph databases handle hierarchical data relationships more effectively. 

Among the  many graphical databases that are available, we choose  to focus on the Neo4j's database and its query language Cypher\cite{francisCypherEvolvingQuery2018}. The \ac{CQL} is a well established query language specifically designed for graph structures. There is a body of research focused on using LLMs to map natural language to Cypher code \cite{hornsteinerRealTimeTexttoCypherQuery2024,mandilaraDecodingMysteryHow2025,ozsoyText2CypherBridgingNatural2024,zhongSyntheT2CGeneratingSynthetic2025,tranRobustTexttoCypherUsing2024} and, indeed, the majority of current LLMs exhibit some exposure to \ac{CQL}.

A core objective of this work is to leverage \ac{LLMs} as an interface between humans and databases for retrieving \ac{IoBT} data. The interface is based on a  two  step approach: 1) conversion of natural language questions into database queries and 2) generation of natural language responses from the database output and user’s original questions. Our approach envisions the collection of real-time IoBT data into a database, either on the device or a nearby network device. Once housed in the database, personnel on-site can ask a question of the data via our system using natural language. A \ac{LLM} first translates the question into a Cypher query to poll the database. The returned database output is combined with the original user’s question for processing into a natural language response back to the user. 
This process streamlines \ac{IoBT} data access for users unfamiliar with database query languages.

The use of a \ac{LLM} in the second step allows the relaxation of the \ac{EM} requirement of 
the predicted Cypher output to ground truth, for the Cypher query generation. This relaxation is enabled in cases where a Cypher query extracts the correct information from the database plus ancillary information. Since the correct ``content'' is in the database response, the second \ac{LLM} call has the opportunity to extract the appropriate information to answer the user's question. 

Our approach investigates some of the current state-of-the-art models that are small enough to be used on-device in the field for answering questions regarding the data and/or the environment. In this way we provide a natural language interface to the user without reliance of cloud resources and knowledge of the query language. 

Key contributions of this work include: 
\begin{itemize}
    \item The design, implementation and evaluation of a pipeline enabling natural language querying of dynamic databases in settings requiring on-device computing;
\item The assessment of the ability of  \ac{LLMs} to generate Cypher code to extract data;
\item The investigation of the capabilities of LLMs to reformulate database answers to queries into natural language sentences;
\item The design of an experimental evaluation framework to assess LLMs' interactions with graphical databases. 
\end{itemize}

Our approach  uses smaller models to generate Cypher queries by only prompting it with information regarding the database schema and one example of a query. Furthermore, we evaluate using a zero-shot modality to align with a scenario of constrained resources as in \ac{IoBT}, e.g. device power limitations.

\section{Related Work}\label{sec:Related_Work}

Natural language querying of databases is a well-established area of investigation in the research literature. Bukhari \textit{et al.} \cite{bukhari_frameworks_2021}, for example, published a literature survey on the topic where, of the 47 frameworks reviewed, they determined 70\% focus on \ac{SQL}, the other 30\% focus on \ac{NoSQL} specific languages, and of these, only 10\% utilize Cypher. Given the increasing use of graph databases, the authors ``urge'' researchers to focus on the development of frameworks for natural language to \ac{NoSQL} for graph databases, such as Cypher, a gap that we address in this work. 

With the rise of \ac{LLMs}, their use in NLP for generating database queries has come to the forefront  \cite{lamas_dsl-xpert_2024, m_mosthaf_natural_2024, bogin_leveraging_2024, pimparkhede_doccgen_2024}. Their use for generating \ac{SQL} queries is actively being researched. In contrast, while research into using LLMs with NoSQL databases is also underway, it has historically been less widespread. 

Recently, research has shifted towards utilizing LLMs with knowledge graphs for knowledge extraction \cite{mandilaraDecodingMysteryHow2025}. Many knowledge graph representations are stored in a NoSQL database like Neo4j, which employs Cypher \cite{francisCypherEvolvingQuery2018}. The Cypher query language has significant complexity, and is challenging for users to learn. This has fueled further research into using LLMs for Cypher query generation, particularly for non-experts seeking to extract data from databases.  

Hornsteiner \textit{et al.} created a chat-based framework utilizing three independent \ac{LLMs} to allow the user to interact with a Cypher database in natural language\cite{hornsteinerRealTimeTexttoCypherQuery2024}. In this framework the user asks a question, which, if it cannot be answered by the chat history, will be sent to a \textit{``Chat LLM''} to generate a Cypher query.  The Cypher query is then sent back to the user for evaluation of errors and, if approved,  it is executed.  While this work has achat functionality, it requires the user to make a decision on the quality of the Cypher query, creating a barrier for non-expert users. In our framework, the use of Cypher queries is entirely abstracted from the user, eliminating the need for any prior familiarity with the query language.   

Along a similar line of research, Mandilara \textit{et al.} define a framework to systematically assess the efficiency of \ac{LLMs} in natural-language-to-Cypher query conversion \cite{mandilaraDecodingMysteryHow2025}. This framework introduces a metric designed to enable comparison of \ac{LLMs} as well as schema-aware prompt engineering methods. While related, their focus is on using the LLMs for only text to Cypher queries, whereas we use the LLMs also for an additional step which summarizes query outputs in natural language. In this sense, our approach to boosting the accuracy of question answering from a graphical database is orthogonal to theirs. They focus on optimizing the Cypher query generation, whereas we propose a novel pipeline that uses two LLMs in sequence and where the role of the second LLM is to mitigate the sub-optimality of the Cypher queries generated by the first one. Moreover, our focus is on tactical environments with constrained resources, aligned with the capabilities of edge devices such as small laptops, tablets, or nodes equipped with hardware accelerators \cite{ngCollaborativeInferenceResourceConstrained2024}. 
In contrast, their focus on selecting the most reliable LLMs for knowledge-driven AI, typically results in models exceeding \ac{IoBT} resource limits.

More complex frameworks that utilize fine-tuning of \ac{LLMs} have been designed to improve \ac{EM} scores. Tran \textit{et al.} \cite{tranRobustTexttoCypherUsing2024} developed a Robust Text-to-Cypher framework using a combination of BERT, GraphSAGE, and a T5 Transformer (CoBGT) model. With this extensive framework the authors observe increases in performance of the \ac{EM} metric of 39.04\% percent over the T5 transformer and 35.81\% percent over GPT2. This framework also has an impressive \ac{EM} value of 87.1\%\cite{tranRobustTexttoCypherUsing2024}. The authors discuss the lack of quality (question,Cypher query) pairs for training data, a drawback also identified by others \cite{tranRobustTexttoCypherUsing2024,zhongSyntheT2CGeneratingSynthetic2025,mandilaraDecodingMysteryHow2025,ozsoyText2CypherBridgingNatural2024} and partially addressed by the publicly available ``Text2Cypher'' datasets \cite{ozsoyText2CypherBridgingNatural2024}. Zhong \textit{et al.}, instead, focus on synthetic dataset generation for supervised fine-tuning. Their methodology enhances output accuracy as much as 12\% with the 8 billion parameter Llama 3-instruct model\cite{zhongSyntheT2CGeneratingSynthetic2025}. 

These works address only one role of LLMs in our system, namely, translating natural language into Cypher queries. In contrast, we also evaluate LLMs for their ability to summarize the query results in natural language. Furthermore, we adopt a zero-shot approach, avoiding fine-tuning, as collecting training data and securing the necessary computational resources may be impractical in resource-constrained tactical environments.

\section{Approach}
\label{sec:appr}
In this section we describe our approach to enable a user to query a graphical database in  natural language, without any knowledge of the database query language.

\begin{figure}[htbp]
\centerline{\includegraphics[width=0.65\columnwidth]{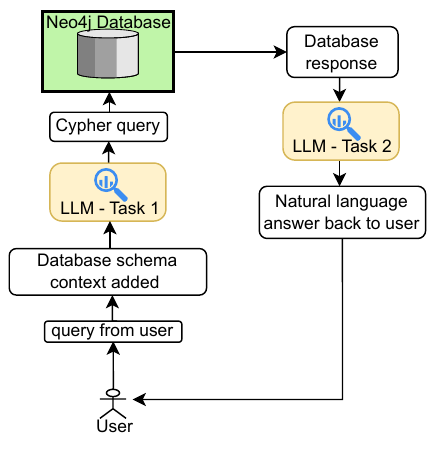}}
\caption{Database  natural language interaction workflow.}
\label{fig:Workflow}
\end{figure}

The workflow we propose, depicted in Fig. \ref{fig:Workflow},  contains two main tasks:
\begin{itemize}
\item \textbf{Task 1:} Translation of a natural language user question into a Cypher query.
\item \textbf{Task 2:} Summarization of  the results of the Cypher query into an answer in  natural language.
\end{itemize}

We propose to employ an LLM in both tasks without requiring them to be the same.

The process flow we envision  starts with a natural language question from a user being  fed into the system. The input to the LLM in Task 1 consists of this question plus some additional context. The output of this task is a Cypher query which, when run,  should extract from the database the data required to answer the question. 

Subsequently, a new prompt is generated including the database output along with the original question and a request to summarize. This prompt is fed to the LLM employed in Task 2 with the goal of providing the correct answer to the user in natural language. 

\begin{figure}[htbp]
\centerline{\includegraphics[width=0.98\columnwidth]{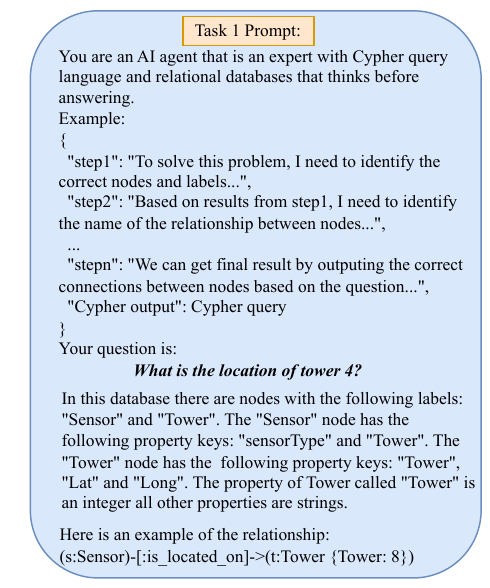}}
\caption{Example prompt used for Task 1.}
\label{fig:Task_1_Prompt}
\end{figure}

An example of a prompt for Task 1 is shown in Fig.~\ref{fig:Task_1_Prompt}. The bold font indicates an example of a question provided by a user. It is preceded by a preamble prompting the LLM to use its Cypher skills and a description of steps to guide its reasoning in producing the Cypher code. The question is then followed by a description of the graphical database schema and a relationship in Cypher code. The example provided in  Fig. \ref{fig:Task_1_Prompt} regards the relationship between a sensor $s$ and the tower on which it is located.

Following our running example more specifically, the output of Task 1 would ideally be a Cypher query allowing to extract the correct data from the database, i.e.:

\noindent \texttt{MATCH (t:Tower {Tower: 4}) RETURN t.Lat AS Lat, t.Long AS Long}. 

Such query can then be included in an API call to the Neo4j database returning: 

\noindent \texttt{[$<$Record Lat=32.58088351 Long = -106.753 3307$>$]}. 

\begin{figure}[htbp]
\centerline{\includegraphics[width=0.90\columnwidth]{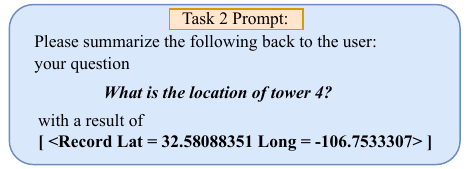}}
\caption{Example prompt used for Task 2.}
\label{fig:Task_2_Prompt}
\end{figure}

Given this, the input prompt for the Task 2 \ac{LLM} would be as shown in Fig.~\ref{fig:Task_2_Prompt}.

A correct output of Task 2 for this case would be a sentence similar to: 

\noindent \emph{The location of Tower 4 is at 32.58088351° latitude and -106.7533307° longitude.}

We note that this example is an illustration of actual results obtained using the Llama3.1:8b model within our workflow.

\section{Experimental design}

We now outline the key components of the experimental evaluation framework we designed to assess the use of different LLMs within the context of our workflow.

\subsection{IoBT Database}
The database was set up to represent the publicly available US Army's \ac{MSA} system, a set of 13 towers within 3 miles of each other.  Each tower has multiple sensors attached. These sensors are primarily used to collect the weather conditions of the local area and include: temperature, humidity, wind speed, precipitation, barometric pressure, soil conditions and network conditions. The sensors are networked together and create a real time \ac{IoT} with the capability of streaming data. A user can capture the live stream of data for storage in raw form or process and store it as consumer-ready information objects.  The current set of experiments carried out in this work involved probing the database representing the tower and sensor setup of the \ac{MSA} system. The extent of the database contains 135 nodes and 121 relationships, along with 11 property types. An example the schema can be seen in the second part of the prompt in Fig. \ref{fig:Task_1_Prompt}. 

\subsection{Questions}

To evaluate our architecture’s question-answering performance, we generate several natural language questions, as shown in   Table~\ref{tab:Questions}. Note that one of the questions is a trick question asking about tower 22, in that tower 22 does not exist.

These questions were selected to exemplify different reasoning depths required to find the  answer  as well as to probe different parts of the database schema.

\begin{table}[htbp]
\caption{Seven questions about the tower and sensor setup for the \ac{MSA} system.}
\begin{center}
\begin{tabular}{c}
\toprule
\textbf{\thead{Questions}}\\ 
\midrule
What sensors are attached to tower 0?\\	
What sensors are attached to tower 12?\\	
What sensors are attached to tower 22?\\	
What is the location of tower 4?\\	
Which towers are closest to each other?	\\
How many sensors are on tower 8?\\
How many towers are there?	\\
\bottomrule
\end{tabular}
\label{tab:Questions}
\end{center}
\end{table}

As described in Section \ref{sec:appr}, the prompt to Task 1 inlcudes, in addition to the user's question, added context for the \ac{LLM} on how to behave and the database schema. The prompt, as shown in Fig. \ref{fig:Task_1_Prompt}, is the result of prompt engineering efforts aimed at optimizing the outputs of  the models employed  in Task 1. All models in Task 1 received the same prompt for a given question.

\subsection{Metrics}

We now provide both an intuitive and a formal description of 
the metrics  we use to evaluate the performance of the LLMs in the workflow. For Task 1 we consider the following: 
\begin{itemize}
    \item \textbf{Exact Match (\ac{EM}) Score:} 
    Measures the number of exact matches between an expert base Cypher query (ground truth) and the Cypher queries produced by the LLM, normalized over the total number of predicted queries.
    \item \textbf{Content Score:} Consists of the number of Cypher queries that return the information needed to answer the user's question when fed into the database,   normalized over the total number of predicted queries. This  can be  verified through parsing of the returned data. 
    \item \textbf{Content Length:} Content Length measures the string length of a correct database response: \ac{EM}s yield the shortest content length, while sub-optimal Cypher queries that still contain the correct information produce longer content lengths.
    \item\textbf{Misinformation Score:} The number of Cypher queries that are syntactically correct but semantically wrong, normalized over the total number of predicted queries. An example of such a query is one that retrieves information about tower 8, even though the user asked about tower 3, perhaps due to confusion caused by a prompt example that referenced tower 8, as in Fig. \ref{fig:Task_1_Prompt}.
\end{itemize}

More formally,  the \ac{EM} score is calculated following Tran \textit{et al} \cite{tranRobustTexttoCypherUsing2024}.   For a single question,  each LLM-generated query $\hat{Y}$ that is an \ac{EM} with the ground truth query  $Y$, gets a value, denoted with $Score_{LF}$,  of 1,  and 0 otherwise. 

The \ac{EM} score oven $N$ questions is then calculated as:
\begin{equation}
  EM = \frac{1}{N}\sum_{n=1}^NScore_{LF}(\hat{Y}_n,Y_n)
  \label{eq:EM_Equation}
\end{equation}

Due to the nature of \ac{LLMs}, predicted Cypher queries are not always \ac{EM}s. They can have a correct syntax and still produce sub-optimal responses containing correct data plus information unrelated to the question. 
To account for this type of result, we utilize the Content Score. In fact, the  variability in the amount of extra information makes it difficult to automate the process to programmatically remove it.  Additionally, the database output is not user-friendly. In such instances, the role of the LLM in Task 2 becomes especially critical, as it must distinguish relevant information from unrelated or ancillary data and respond to the user in natural language.

A key point proved here is that a team of \ac{LLMs} can be utilized to improve the output to the user without specialized training or fine-tuning. This is critically important for deployment on edge devices in the \ac{IoBT} where additional training may be unfeasible due to computational constraints. 

The \ac{LLM} employed in  Task 2 is also graded on one additional metric, the Output Score: 
\begin{itemize}
\item \textbf{Output Score}: This score represents the number of correct responses for Task 2 for a given database result as input.  The output of the database can be in one of the following four possible cases: 1) It contains the data necessary to answer the user's question correctly; 2) It is the empty list ``[]'', meaning that the semantic relationship in the predicted Cypher query is incorrect or the searched data does not exist in the database; 3) It is ``$nan$'', meaning that a syntax error was returned caused by an incorrect Cypher query prediction; 4) There is a returned response, but it does not contain the data required to answer the question. This occurs when the predicted query is structurally valid but fails semantically, resulting in the extraction of incorrect data.
\end{itemize}
The overall workflow is graded by an Absolute Score:
\begin{itemize}
    \item \textbf{Absolute Score:} This global metric is the number of  correct responses produced by the system, normalized over the total number of inputs.
\end{itemize}

\section{Results and Discussion}

We evaluate multiple \ac{LLMs} in this workflow to understand their capabilities and limitations as it pertains to generation of Cypher code (Task 1) and reformulating answers to database queries in natural language (Task 2). 

Given our focus on edge devices, we only consider  \ac{LLMs} of medium size, as defined in \cite{minaee_large_2024}, between 1b and 10b parameters. We test models near both ends of this size range to ensure compatibility with edge devices with different capabilities in terms of memory and power. 

The software utilized for these experiments includes: Neo4j Database Desktop version 1.6.1, the Ollama LLM platform version 0.5.4, and Python version 3.12.  The models used with Ollama include: Gemma2 (2b parameter), Llama3.2 (3b parameter), Llama3.1 (8b parameter) and Deepseek-coder (6.7b parameter).  
All experiments were run on a Dell XPS 15 Laptop with a 13th Generation Intel i9 2.60 GHz processor, 32 Gb of memory and Nvidia GeForce TRX 4060 Laptop GPU with 8 Gb memory.

Given the seven questions shown in Table \ref{tab:Questions}, the outputs produced by the workflow are assessed for EM, Content, Output and Absolute Scores with results shown in Table \ref{tab:EMScores}.  In this study, we used the same LLM for both Task 1 and Task 2. Given the nature of our tasks, all models were run with the temperature set to zero. 

All of the LLMs employed in Task 1 achieved high Content Scores (above 71\%), indicating that their answers to the user's questions (i.e., Cypher queries) produce the correct  database result most of the time. However, the queries often were not exact matches (see Table \ref{tab:EMScores}, first column). The job of the LLM in Task 2 is to extract that content from the database result and return the correct response to the user. For all the models employed in Task 2 the Output Score is above  71\% and is, above the Content Score for all models except for Deepseek-coder:6.7b,  suggesting that the task of summarizing data output in natural language is easier than translating the user's question into Cypher queries. 

\begin{table}[htbp]
\caption{Performance for the four LLMs evaluated using 7 questions.}
\begin{center}
\resizebox{\columnwidth}{!}{
\begin{tabular}{ |c|c|c|c|c| }
\hline
\textbf{\thead{Model}} 	&	\textbf{\thead{\ac{EM} score\\ Task 1}} 	&	\textbf{\thead{Content Score\\ Task 1}} 	&	\textbf{\thead{Output Score\\ Task 2}} 	&	\textbf{\thead{Absolute\\ Score}} 
 \\
\hline
Gemma2:2b	&	14.3\%	&	71.4\%	&	85.7\%	&	57.1\%	\\
Llama3.2:3b	&	28.6\%	&	71.4\%	&	100\%	&	71.4\%	\\
Llama3.1:8b	&	42.9\%	&	85.7\%	&	100\%	&	85.7\%	\\
Deepseek-coder:6.7b	&	28.6\%	&	85.7\%	&	71.4\%	&	57.1\%	\\
\hline
\end{tabular}
}
\label{tab:EMScores}
\end{center}
\end{table}

In general, each model produced differing Cypher queries for the same input, which  resulted in varying outputs from the database. We illustrate this in Table \ref{tab:Content_Length} for the question ``What is the location of tower 4?''. While all answers contained the correct information,  only Llama3.1:8b has an \ac{EM}, resulting in the minimum  Content Length of 44 characters, depicted in Fig. \ref{fig:Task_2_Prompt}. The Llama3.2:3b model produced  the longest Content Length of 2171 characters. The variability reported in Table \ref{tab:Content_Length}, and the Content Scores in Tables \ref{tab:EMScores} and \ref{tab:scores_77_questions}, illustrate both the  variability in the output generated by the predicted Cypher queries, which often includes superfluos data, as well as 
the LLMs’ capacity to extract the meaningful information from noisy or unrelated input.

\begin{table}[htbp]
\caption{Content Length variation from returned database queries.}
\begin{center}
\resizebox{\columnwidth}{!}{
\begin{tabular}{ |c|c|c| }
\hline
\textbf{\thead{Model}} 	&	\textbf{\thead{User Question}} 	&	\textbf{\thead{Content Length\\ of DB output}} 	\\
\hline
Gemma2:2b	&	What is the location of tower 4?	&	167	\\
Llama3.2:3b	&	What is the location of tower 4?	&	2171	\\
Llama3.1:8b	&	What is the location of tower 4?	&	44	\\
Deepseek-coder:6.7b	&	What is the location of tower 4?	&	624	\\

\hline
\end{tabular}
}
\label{tab:Content_Length}
\end{center}
\end{table}
To assess the robustness of our system to different ways of asking the same question we used the  Llama3.1:8b  model to rephrase each of the original questions 10 times and we manually checked them for correctness. EM, Content, Output and Absolute Scores for the 77 questions are shown in Table \ref{tab:scores_77_questions}. As expected, most scores decline; however, the key trends persist: Content Scores remain higher than EM Scores, and Output Scores exceed Content Scores. This supports our hypothesis that exact match (EM) may be an overly stringent metric, as the correct information is often present despite being embedded in superfluous content, and LLMs are generally effective at extracting and presenting this information in natural language.
\begin{table}[htbp]
\caption{Performance for the four LLMs evaluated using the 77 questions. }
\begin{center}
\resizebox{\columnwidth}{!}{
\begin{tabular}{|l|c|c|c|c|}
\hline
\textbf{\thead{Model}} 	&	\textbf{\thead{\ac{EM} score\\ Task 1}} 	&	\textbf{\thead{Content Score\\ Task 1}} 	&	\textbf{\thead{Output Score\\ Task 2}} &	\textbf{\thead{Absolute\\ Score}}\\
\hline
Gemma2:2b & 15.6\% & 51.9\% & 76.6\% & 33.8\%\\
Llama3.2:3b & 15.6\% & 57.1\% & 85.7\% & 44.2\% \\
Llama3.1:8b & 37.7\% & 61.0\% & 96.1\% & 57.1\% \\
Deepseek-coder:6.7b & 20.78\% & 42.9\% & 74.0\% & 31.2\% \\
\hline
\end{tabular}
}
\label{tab:scores_77_questions}
\end{center}
\end{table}
We can further evaluate our workflow by comparing the Absolute Score (EM Only), corresponding to only EM cases, with the Absolute Score as previously defined (when the Content Score is considered for Task 1),  as shown in Table \ref{tab:absolute_score_compare_77_questions}.  We observe a marked improvement in the Absolute Scores, with Llama3.1:8b scoring the highest with 57.1\%  and the other models more than doubling their Absolute Score.
\begin{table}[htbp]
\caption{Absolute Score for EM only queries vs Absolute Score for the 77 questions.}
\begin{center}
\begin{tabular}{|l|c|c|}
\hline
\textbf{\thead{Model}} 	&	\textbf{\thead{Absolute Score\\ (EM Only)}} 	&	\textbf{\thead{Absolute\\ Score}}\\
\hline
Gemma2:2b & 14.3\% & 33.8\% \\
Llama3.2:3b & 14.3\% & 44.2\% \\
Llama3.1:8b & 37.7\% & 57.1\% \\
Deepseek-coder:6.7b & 14.3\% & 31.2\% \\
\hline
\end{tabular}
\label{tab:absolute_score_compare_77_questions}
\end{center}
\end{table}
For all models, incorrect answers arise when results from Task 1 contain syntax, semantic, or misinformation errors, or when Task 2 fails to extract available content. The misinformation error is especially subtle, as it feeds incorrect information to Task 2 which is then  propagated  to the user. This error, quantified as the Misinformation Score, measures the frequency of misinformed answers to the users' question (see Table  \ref{tab:misinformation_score_77_questions}). To be noted is Llama3.1:8b's extremely low rate of 5.2\%.
\begin{table}[htbp]
\caption{Misinformation Scores for the 77 questions.}
\begin{center}
\begin{tabular}{lc}
\toprule
\textbf{\thead{Model}} 	&	\textbf{Misinformation Score} \\
\midrule
Gemma2:2b & 35.1\%  \\
Llama3.2:3b & 7.8\%  \\
Llama3.1:8b & 5.2\%  \\
Deepseek-coder:6.7b & 26.0\%  \\
\bottomrule
\end{tabular}
\label{tab:misinformation_score_77_questions}
\end{center}
\end{table}
Manual analysis revealed consistent errors in Cypher queries stemmed from incorrect tower numbers (often using the engineered prompt's example instead of the user's input), not misunderstandings of the question itself. This insight can guide future refinements of the Task 1 prompt.

Overall, we note that, for all of the assessments and metrics, Llama3.1 is the top model in this workflow scoring the highest Absolute Score (57.1\%) and the lowest Misinformation Score (5.2\%)
make it a promising candidate for further testing to enable its deployment on edge devices. 

\section{Future Work}
Our future research agenda includes: testing our workflow under dynamic conditions with  live stream data from the \ac{MSA} system; the investigation of other  models with reasoning capabilities (e.g., Deepseek-r1 and Gemma 3); and investigate  in-context learning as well as pairing this workflow with light weight model fine-tuning to increase overall performance.


\bibliographystyle{IEEEtran}
\bibliography{references}

\end{document}